# Data-driven development of cycle prediction models for lithium metal batteries using multi modal mining

*Jaewoong Lee[1,†], Junhee Woo[2,†], Sejin Kim[1], Cinthya Paulina[2], Hyunmin Park[2], Hee-Tak Kim[1,*], Steve Park[,2*] and Jihan Kim[1,*]*

[1] Department of Chemical and Biomolecular Engineering, Korea Advanced Institute of Science and Technology, Daejeon, 34141, Republic of Korea

[2] Department of Materials Science and Engineering, Korea Advanced Institute of Science and Technology, Daejeon, 34141, Republic of Korea

† These authors contributed equally: J. Lee, J. Woo

*: Corresponding author

Corresponding author Email: Jihankim@kaist.ac.kr (Jihan Kim), stevepark@kaist.ac.kr (Steve Park), heetak.kim@kaist.ac.kr (Hee-Tak Kim)

**Abstract**

Recent advances in data-driven research have shown great potential in understanding the intricate relationships between materials and their performances. Herein, we introduce a novel multi modal data-driven approach employing an Automatic Battery data Collector (ABC) that integrates a large language model (LLM) with an automatic graph mining tool, Material Graph Digitizer (MatGD). This platform enables state-of-the-art accurate extraction of battery material data and cyclability performance metrics from diverse textual and graphical data sources. From the database derived through the ABC platform, we developed machine learning models that can accurately predict the capacity and stability of lithium metal batteries, which is the first-ever model developed to achieve such predictions. Our models were also experimentally validated, confirming practical applicability and reliability of our data-driven approach.

**INTRODUCTION**

Lithium metal batteries (LMBs) are a promising next-generation device that can achieve high capacity using lithium metal as an anode due to its exceptionally low density (0.534 g cm$^{-3}$), high theoretical capacity (~3860 mAh g$^{-1}$), and low electric potential (-3.04 V compared to the standard hydrogen electrode). Unfortunately, the high reactivity of lithium metal poses an obstacle to the long-term cyclability of LMBs and research is being conducted in various fields to overcome this issue[1-4]. Nevertheless, due to delayed performance feedback caused by the extended time required to measure the cycle life of batteries, an early prediction of cycle life from initial performance remains a key challenge that impedes the development-validation cycle of materials and manufacturing processes[5-8]. Overall, the performance of a battery depends heavily on the structure and composition of the materials[9-13]. While the individual characteristics of each component are important, a comprehensive understanding of how these components interact with each other is crucial[14] as the cycle life of a battery cell heavily depends on these complex correlations.

Recently, data-driven approach method has become a powerful paradigm to understand complex relationships between different parameters and to discover patterns that can be applied to enhance prediction from numerous experimental data sets[15]. Notably, the robustness of data-driven approaches has been demonstrated in several studies using machine learning with experimental battery datasets (e.g. predicting battery lifetimes[16], identifying degradation patterns from electrochemical impedance spectroscopy[17], designing high-performance electrolyte for lithium metal anode[18], analyzing lithium electrode potential[19], and developing fast-charging protocols[20]). Moreover, with the advancement of data mining technologies, efforts have been made to extract experimental information from battery literature using natural language processing (NLP)[21] and transformer models[22,23].

However, previous studies controlled for only one or two variables (e.g., electrolytes, and charge/discharge condition). Therefore, these studies lack sufficient information to discern a comprehensive effect of different components on the battery performance. Additionally, previous mining research focused not on the entire battery cells but rather on the characteristics of individual battery components. Moreover, these studies were limited by the small number of entities considered and did not extract quantitative information such as concentrations or ratios. Furthermore, the absence of automatic graph mining tools made it difficult to obtain performance data from graphs, such as specific capacity and cycle stability. These limitations make it challenging to fully explore the material-performance relationship and utilize it for machine learning studies. Therefore, to unlock the full potential of innovative battery technologies, it is essential to employ a new method that leverages comprehensive data to understand the entire battery cycle.

Here, we present a novel multi modal data-driven approach utilizing an Automatic Battery data Collector (ABC) platform consisting of a large language model[24-26] (LLM) and an automatic graph mining tool[27] (Material Graph Digitizer, MatGD). We automatically extracted comprehensive information regarding battery materials with 96.4% accuracy, which is the state-of-the-art performance known to date, encompassing a total of 29 entities, the largest number of entities ever achieved. Additionally, for the first time, successfully mined cyclability performance data were directly extracted from cycle graphs. Using this uniquely extensive dataset, we developed novel machine learning models capable of predicting initial capacity, target cycle capacity, and stability at target cycles. Our models are the first of their kind not only in their predictive capability but also in their experimental validations.

# Results

## Overall workflow

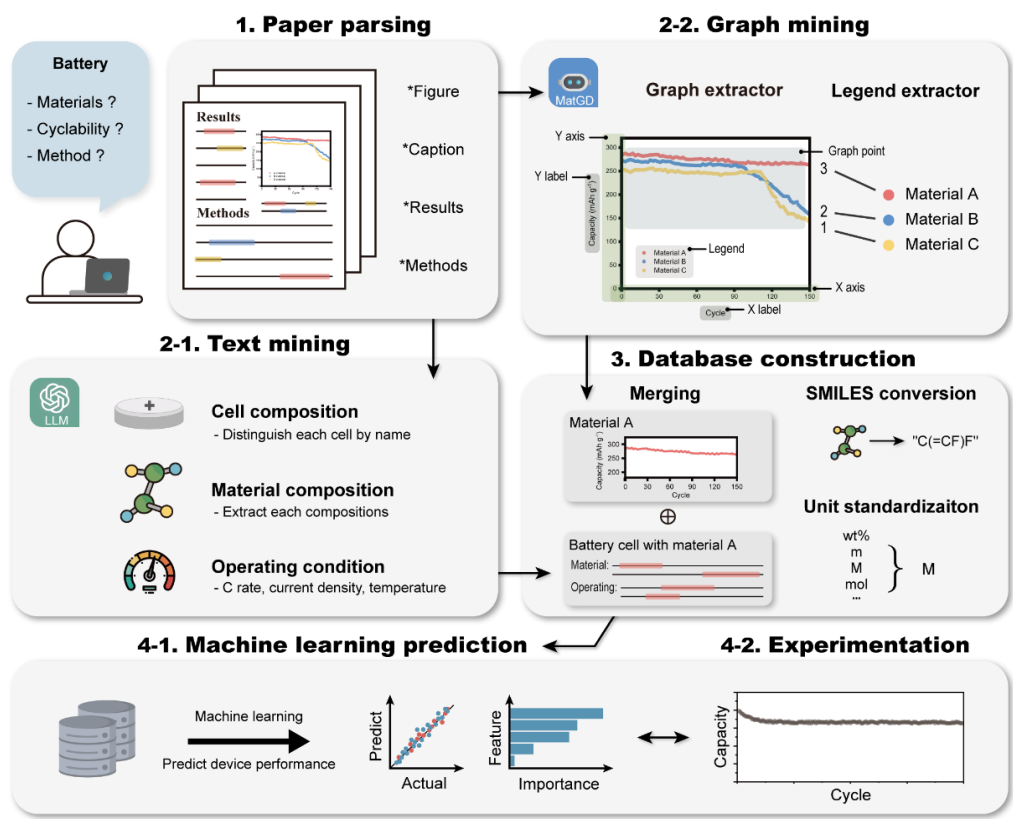

**Fig. 1: Machine learning-assisted data-driven approach in battery science**
A schematic image of the entire process in our data-driven approach. The ABC automatically extracts material, cyclability, and method data from the literature. Machine learning models predict cycle performance of LMBs utilizing the database, and their validity is confirmed through experimental verification.

The overall workflow of our research is schematically illustrated in Fig. 1. We utilized the ABC platform for data mining and database construction, which was subsequently applied in machine learning applications., Overall, the ABC platform consists of four distinct agents responsible for parsing, text mining, graph mining, and database construction.

To extract data, a parsing agent first crawled a total of 6,444 LMB-related papers using the Elsevier Scopus API (Supplementary Fig. 1). It then parsed the downloaded papers into caption, result, method, and graph sections. The text mining agent processes the caption, result, and method sections to extract information regarding battery materials and operating conditions. This is further divided into three extraction processes: 1) cell composition, 2) material compositions, and 3) operating conditions. To collect cyclability data from the figures, we used MatGD, an automatic graph digitization tool developed by our research group. The graph mining agent receives the metadata (DOI, figure number, graph label) of cycle graphs identified by cycle graph classification stage of text mining and uses MatGD to automatically collect information from these graphs. This includes extracting each data point, X-Y axes, labels, and the graph legend. Finally, the database construction agent merges the data obtained from the text mining and graph mining procedures to construct a complete battery database. Standardization procedures (e.g. SMILES conversion, unit conversion) are applied to obtain homogeneous dataset, which can facilitate the machine learning process. Through this end, we constructed a comprehensive database of 8,074 battery cells, incorporating both text and graph data, containing component specifics and cycle-dependent capacity. Further explanations of each process are provided in the respective sections.

# Text mining agent

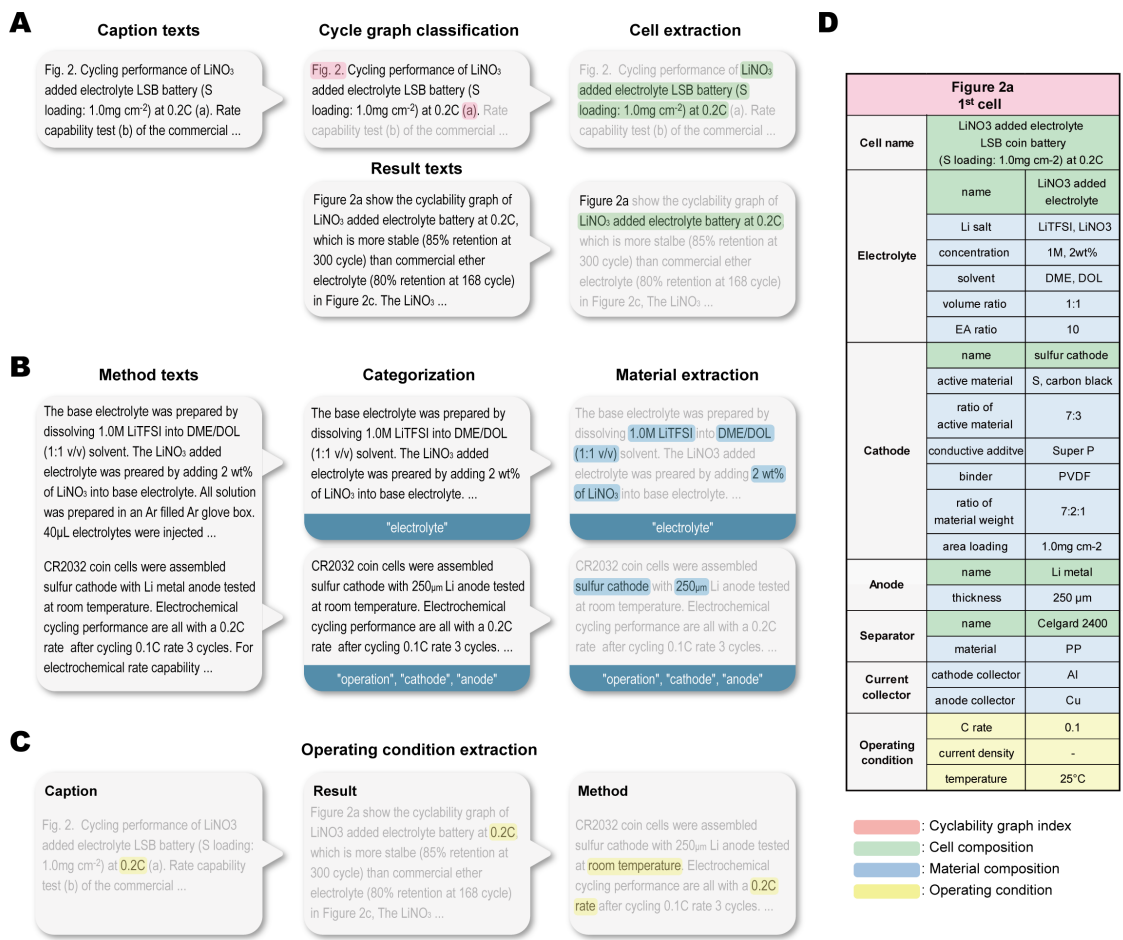

**Fig. 2: Example of the text mining procedure.** Overall process of each text extraction process is illustrated: **a** cycle graph metadata and cell composition, **b** material composition, and **c** operating condition. The extraction results are highlighted and the categorization results are written within the gray boxes. **d** An example of the text mining result, with relevant content colorized for each extraction.

To gain a comprehensive understanding about the battery cells, information such as cycle graphs, cell composition, material compositions, and operating conditions are required. However, this information is widely scattered in a given paper across the caption, result, and method paragraphs. To remedy this, the text mining agent integrates multiple LLMs (Supplementary Fig. 2), that focused on specific tasks with different purposes[28], to optimize performance. Among the various LLMs available, we based our architecture on the GPT-4.0 model due to its high accuracy and rapid processing speed. The entire process is automated through the Python code, requiring only 2-4 minutes per paper. This GPT-4.0 based text mining approach enable efficient construction of a comprehensive database.

Fig. 2 illustrates an example of each text mining process with the extracted results highlighted in different colors. First, the text mining agent extracts information regarding the cell composition to distinguish individual battery cells in the graph (Fig. 2a). During the cycle graph classification stage, the agent analyzes the captions to identify which figures are associated with cycle graphs by determining if each caption includes context related to the cycling test. The graph metadata for the cycle graph is then passed on to the cell extraction stage. Using a Python code, the relevant paragraphs within the manuscript associated with the cycle graph metadata are identified and retrieved. Subsequently, the information about the names of cells and components present in the cycle graph, along with the corresponding captions and results, is extracted in the cell extraction stage.

Next, material composition is extracted as illustrated from an example in Fig. 2b. This stage extracts the composition information of materials from the method section. This process is divided into two stages: categorization and material extraction. The categorization stage distinguishes the type of material information contained in the method paragraphs and is further divided into major- and sub-categorization for efficient classification (Supplementary Fig. 3). The major-categorization classifies paragraphs into 'material', 'synthesis', 'operating condition', and 'other' categories. The 'synthesis' category filters out unrelated paragraphs regarding the synthesis of non-battery materials to enhance accuracy. The 'material'-classified paragraphs are sub-categorized based on battery cell components ('cathode', 'anode', 'electrolyte', 'separator', 'current collector', and 'other'), and 'operation' - classified paragraphs are further sub-categorized into 'cycle performance', and 'other' categories. This approach enables efficient categorization of battery information within each paragraph, thereby reducing the GPT token counts and enhancing the accuracy of text mining. The material extraction stage extracts individual composition data of materials based on the names of each cell and component extracted in the cell extraction stage, from the related paragraphs. For example, the electrolyte salts and solvents are identified and stored as numerical values, including their concentration and quantity.

Fig. 2c presents the process of extracting the cyclability operating conditions to accurately compare cyclability performance. Information related to the C-rate, current density, and temperature is identified and extracted from the caption, result, and method paragraphs. This information is most frequently and accurately mentioned in the caption, whereas in the result and method sections, it is often less clearly specified as a range value (e.g. 0.2 — 1.0 C-rate, 25°C — 45°C). Therefore, the operating condition extraction stage was conducted in the order of caption, result, and method paragraphs.

The extracted data for cycle graph metadata (red), cell composition (green), material composition (blue), and operating conditions (yellow) are compiled into a JSON file (Fig. 2d). This resulting data, which includes a total of 29 entities such as material names, property values, and units, serves as the input for the machine learning models.

## Graph mining agent

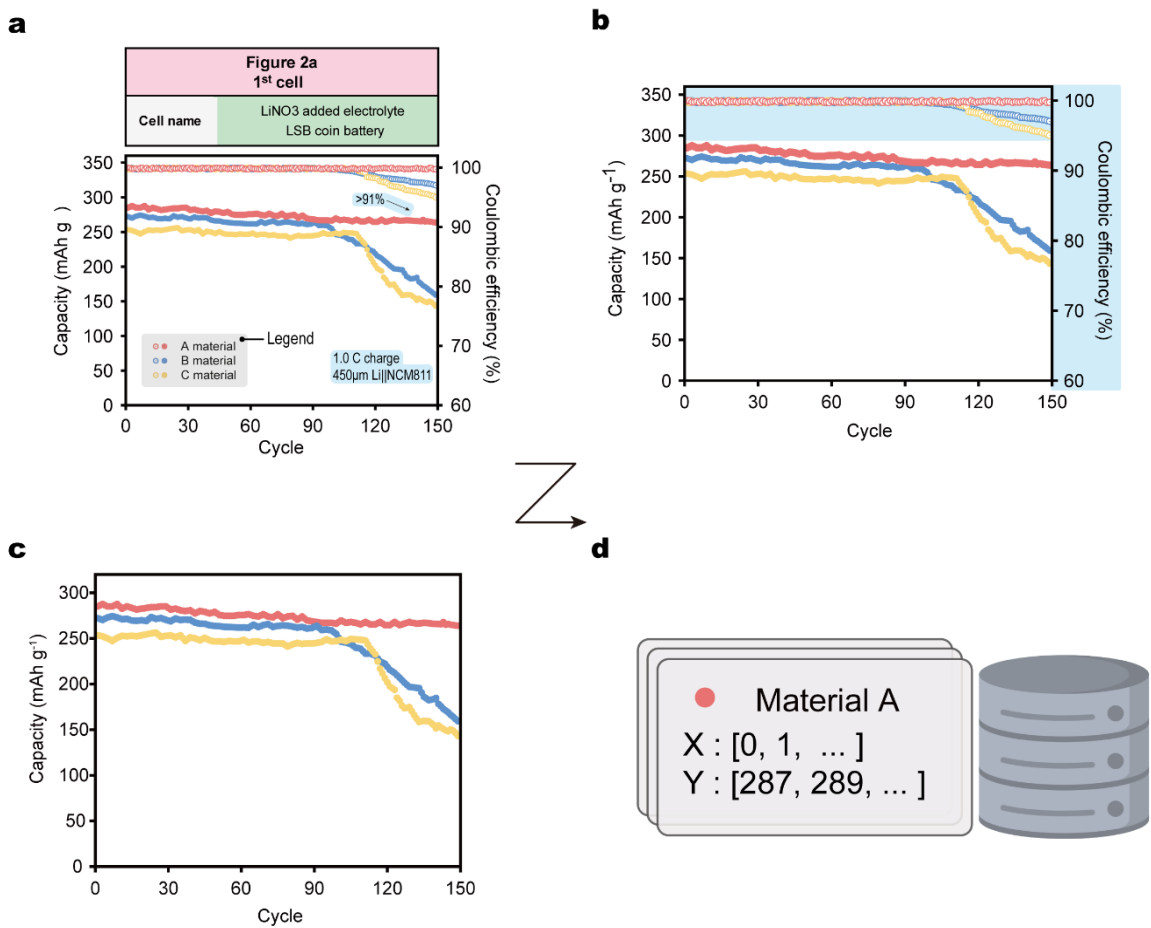

**Fig. 3: Example of the cycle graph mining procedure. a** Decomposition of each element of the cycle graph distinguished by the caption while removing unnecessary parts (highlighted in blue). **b** Detecting and **c** eliminating CE from the cycle graph. **d** Saving the name of the cell in the legend and recording the cycle graph data point by point.

An example that summarizes our graph mining procedure is depicted in Fig. 3. First, 6,682 figures in 3,606 papers classified as a cycle graph in the cycle graph classification stage were downloaded (Fig. 3a) from the Elsevier Scopus API. From these figures, 1,635 graphs could not be extracted using MatGD, our graph digitizing tool due to color-based data separation reasons. Detailed information regarding the classification process is described in the method section. The rest of the cycle graphs were processed using MatGD, which removed text, markers, arrows, and other graph objects, leaving only the data line values (Fig. 3b). Furthermore, the information about the Coulombic efficiency (CE) is present in many of the cycle graphs, which needs to be removed to exclusively extract the cyclability data. The distinct properties of the CE were identified and a rule-based code was developed and used to remove this data, allowing us to obtain only the cyclability data (Fig. 3c). Detailed information regarding this process is provided in the graph mining section of the Methods. Finally, the cyclability data for each battery cell were saved in the JSON format (Fig. 3d).

## Database construction agent

Each piece of the data collected from the text and graph mining agents exists independently without being dependent on a specific battery cell. Therefore, it is important to aggregate all the data into one database. Several data processing stages were implemented to create a battery database that facilitates using the relevant data for machine learning purposes. First, we integrated the information on battery cells obtained through text mining with the cyclability data extracted from graph mining. To enhance data integration, we employed prompt engineering with the GPT-4 model. The inputs for the merging process were the names of battery components extracted through text mining, and the labels of data lines obtained from graph mining. Distinguishing features, such as cathode material or operating conditions, which are pivotal for identifying different battery cells are implicitly included within the labels, ensuring high merging accuracy when aligned with names mentioned in textual data. Second, after integrating all the data, standardization procedure is needed given that many of the naming conventions are different across thousands of different published papers. Specifically, properties such as thickness of lithium anodes, temperature, current density, and loading amount of active material all need to be in the same units. For components composed of small molecules and polymers (such as binders, conductive materials, electrolyte salts, electrolyte solvents, and separators), their names were converted into the SMILES format. Lastly, to unify units involving different dimensions (including the concentration of electrolyte salts and the ratio of electrolyte solvents), additional parameters like molecular weight and density were used for the conversion (Supplementary Table 1).

## Mining Performance for the Extracted Data

| | # of Paper | # of Graph | # of Cell | Evaluation metrics | | | | |
|---|---|---|---|---|---|---|---|---|
| | | | | Accuracy | Precision | Recall | F1 score | |
| **Text Mining** | 3,606 | 6,682 | 15,398 | - | 0.986 | 0.942 | 0.964 | Cell |
| | | | | 0.959 | 0.963 | 0.989 | 0.976 | Cathode |
| | | | | 0.983 | 0.985 | 0.991 | 0.988 | Electrolyte |
| | | | | 0.995 | 1.000 | 0.980 | 0.990 | Anode |
| | | | | 0.968 | 0.958 | 1.000 | 0.979 | Separator |
| | | | | 0.987 | 0.986 | 0.994 | 0.990 | Current Collector |
| | | | | 0.979 | 0.943 | 1.000 | 0.971 | Operating condition |
| **Graph Mining** | 3,044 | 5,047 | 10,242 | - | - | 0.665 | - | |
| **Final Database** | 2,549 | 3,567 | 8,074 | 0.989 | 0.986 | 1.000 | 0.993 | Merging |

**Table 1.** The number of papers, graphs, cell results and accuracy, precision, recall, F1 score of each stage. The results of the text mining are summarized.

The statistics and evaluation metrics of ABC results are shown in Table 1. To analyze the accuracies of the text mining, graph mining, and the database merging steps, we quantify the number of papers, graph, and cell with the evaluation metrics of accuracy, precision, recall, and F1 score. As described in the Text Mining Agent section, the names of each cell undergoing cyclability testing, along with the components such as the cathode, electrolyte, and others, were extracted separately. This approach allowed for the application of evaluation metrics to each individual component.

After removing papers without the cycle graphs, a total of 15,398 battery cells were extracted from 3,606 LMB papers downloaded from the Elsevier journal. From this, a random sample of 384 battery cells from 100 papers were analyzed to measure the performance of our text mining tool. Each component showed a high F1 score, with the current collector achieving 0.990, and even the lowest value, for cell extraction, still being 0.964. The details of other evaluations are presented in Supplementary Tables 4-8.

Due to the high accuracy of the text mining agent, the number of battery cells extracted through the text mining procedure were used as truth values to evaluate the recall values of the graph mining results. Out of 15,398 cells, 10,242 cyclability data were extracted by MatGD, resulting in a recall value of 0.655. Note that this recall value reflects the limitations of MatGD's color-based separation, which may lead to fewer extractions than possible. However, all of the extracted data were manually verified to ensure that only accurate cyclability values were included, thereby building a robust database and ensuring that most of the data used in the database or machine learning models are reliable. Afterwards, for the construction of a refined database, the text and graph mining results were integrated. In the evaluation of 265 matching results from a test set of 100 randomly selected papers, we achieved a very high recall value of 1.000 and an F1 score of 0.993. As a result, a valid database that contains information regarding 8,074 battery cell components and cyclability performance data were constructed using our platform.

# ABC results

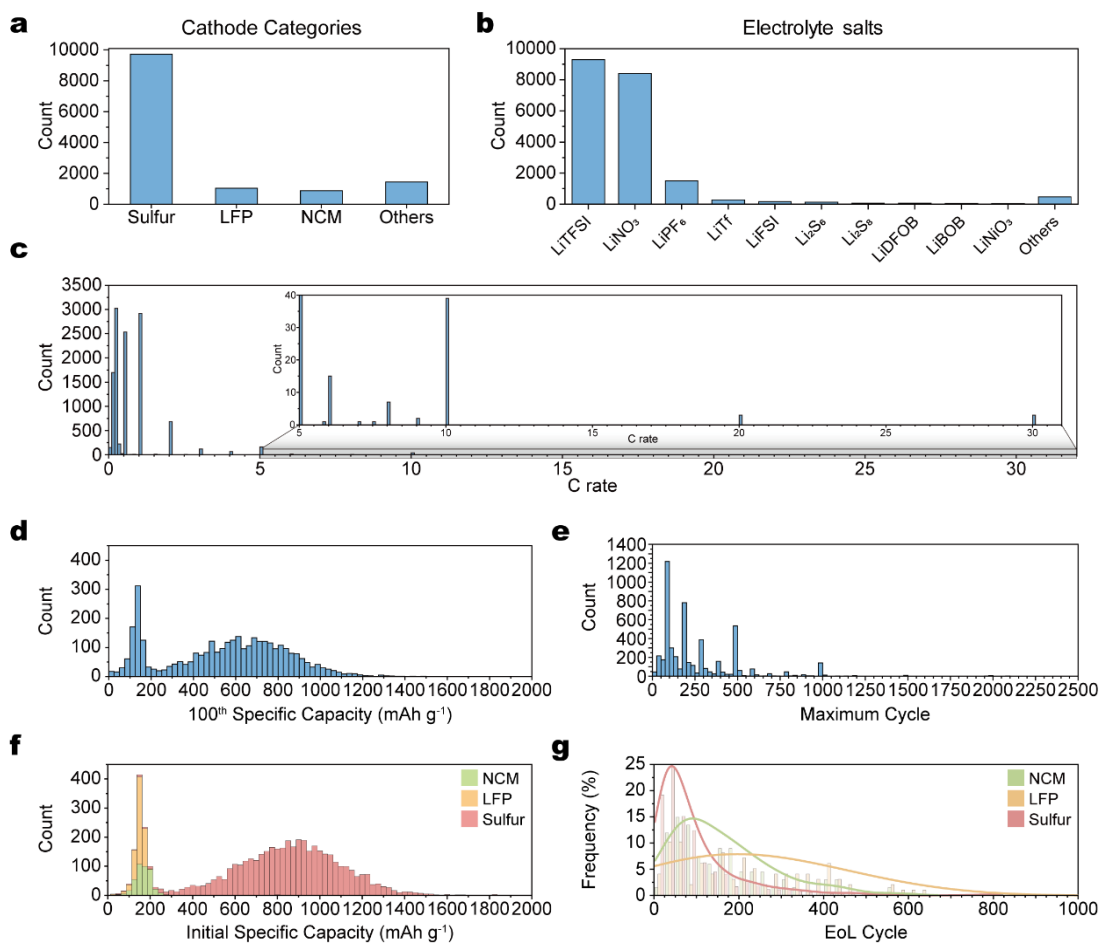

**Fig. 4: Results of mined features and ABC database of LMBs.** Data distribution of material compositions of **a** cathode categories (13,069 cells), **b** electrolyte salts (11,511 cells), and operating conditions of **c** C-rate (11,729 cells) extracted from text mining. Histogram plot of **d** specific capacity at $100^{th}$ (3,469 cells), and **e** maximum cycle number data (5,123 cells) extracted from graph mining. Histogram plot of **f** initial capacity (8,687 cells), and **g** EOL data in the merged database (1,736 cells), with NCM, LFP, and sulfur cathode cells represented in green, orange, and red, respectively.

Fig. 4 illustrates the distribution of data mining results. From the collected data, it can be seen that sulfur is the most extensively studied cathode material, followed by LFP, NCM, and other lithium transition metal oxides (Fig. 4a). According to the data distribution of electrolyte salts, LiTFSI has been the most commonly used, followed by $LiNO_3$ and $LiPF_6$ (Fig. 4b). The most frequently used C-rate conditions are 0.2~0.3 and 1.0~1.1. Detailed distributions of other material compositions and operating conditions can be found in Supplementary Fig. 4-13 and 14, respectively.

Next, the data of LMBs (i.e. specific capacity at the $100^{th}$ cycle and the maximum cycle data) extracted from the graph mining agent is visualized in Fig. 4d and Fig. 4e, The histogram of the $100^{th}$ cycle capacity shows distinct peaks at around 170 and 700 mAh $g^{-1}$, which correspond to different cathode types as detailed in the merged results. The histogram of initial capacity data, which merges text and graph mining results, is plotted in Fig. 4f. It can be seen that the peak values associated with NCM and LFP cathode LMBs are much lower compared to LMBs, which have a much higher theoretical specific capacity at 1675 mAh $g^{-1}$ (Supplementary Fig. 15). Finally, the end of the life (EOL) data is plotted for each of the three cathodes and it can be seen that in general, LFP and NCM cathode LMBs demonstrate better long-term stability than sulfur cathode LMBs (Fig. 4g), which agrees with the notion that that the lithium sulfur battery (LSB) exhibits low stability despite being operated at a relatively low C-rate (Supplementary Fig. 16).

## Machine learning Results

From the data collected in our mining procedure, we looked to enable early prediction of cycle performance using different machine learning models. eSpecifically, these models were designed to accomplish three key tasks: 1) predicting initial capacity from the battery material, 2) predicting capacity at target cycles given the material and initial capacity data, and 3) predicting stability at target cycles from the material. The material composition input features used in the machine learning models are shown in Supplementary Table 9.

### *1. Prediction of initial capacity*

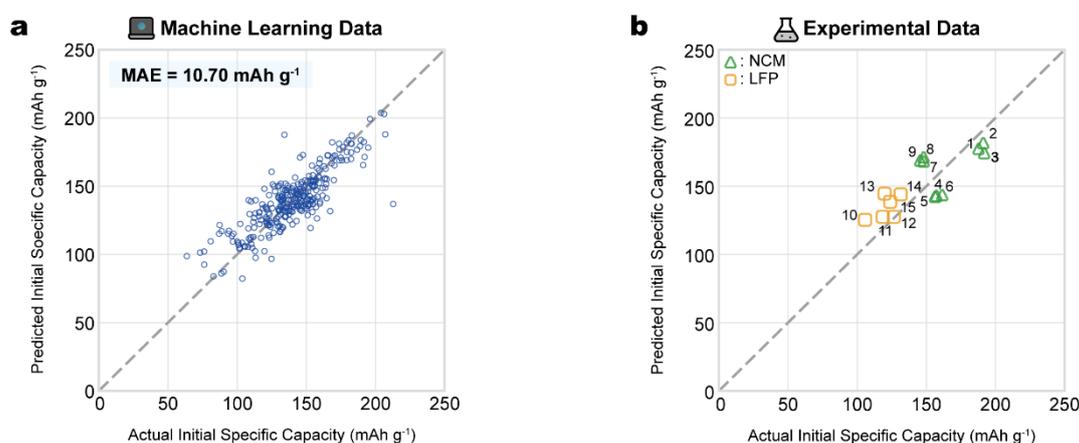

**Fig. 5: Prediction of initial capacity.** Result of initial capacities predicted from the random forest model with **a** the mined database and **b** our own experimental data. Those performances of the batteries are based on LMB with NCM and LFP cathodes., represented as green triangles and orange squares, respectively.

Fig. 5 presents the machine learning results for predicting the initial capacity of NCM, LFP cathode LMBs based on their material compositions (Supplementary Fig. 17) in absence of any cycle performance data. A total of 87 features were used, with 295 data points divided into 206 training data points and the rest used as test data. The random forest (RF) model recorded the highest MAE of 10.70 mAh $g^{-1}$ (Supplementary Table 10). Fig. 5a plots the predicted initial capacity values for each cell data across different folds whereas the actual initial specific capacity was taken from the text mined data. The model was validated using our own experimental data conducted under 15 different LMB conditions, all of which were not used in model training (Fig. 5b, Supplementary Table 14, 15). From Fig. 5b, it can be seen that there is in general good agreement between the predictions made by our machine learning model and the actual experimental data. Moreover, the NCM and LFP are clustered separately, confirming that the model has learned to distinguish the features between two different classes of batteries.

On the other hand, LSBs showed poor performance, with a MAE of 160.99 mAh $g^{-1}$ (Supplementary Fig. 18). For sulfur, the electrochemical reaction behavior with $Li^+$ ion depends on the type and structure of the host and sulfur's incorporation, which in turn affects its utilization. Therefore, detailed material information about the host is crucial for predicting initial capacity. However, due to the diversity and complexity of host materials, we were not able to encode them in machine learning features. Additionally, efforts to train models separately for carbon hosts and non-carbon hosts were made, but these did not significantly enhance accuracy (Supplementary Table 11). Given the high performance observed in NCM and LFP, it can be inferred that a new method to encode the sulfur/host composite must be actualized to accurately predict the initial capacity of LSBs.

*2. Prediction of capacity at target cycle*

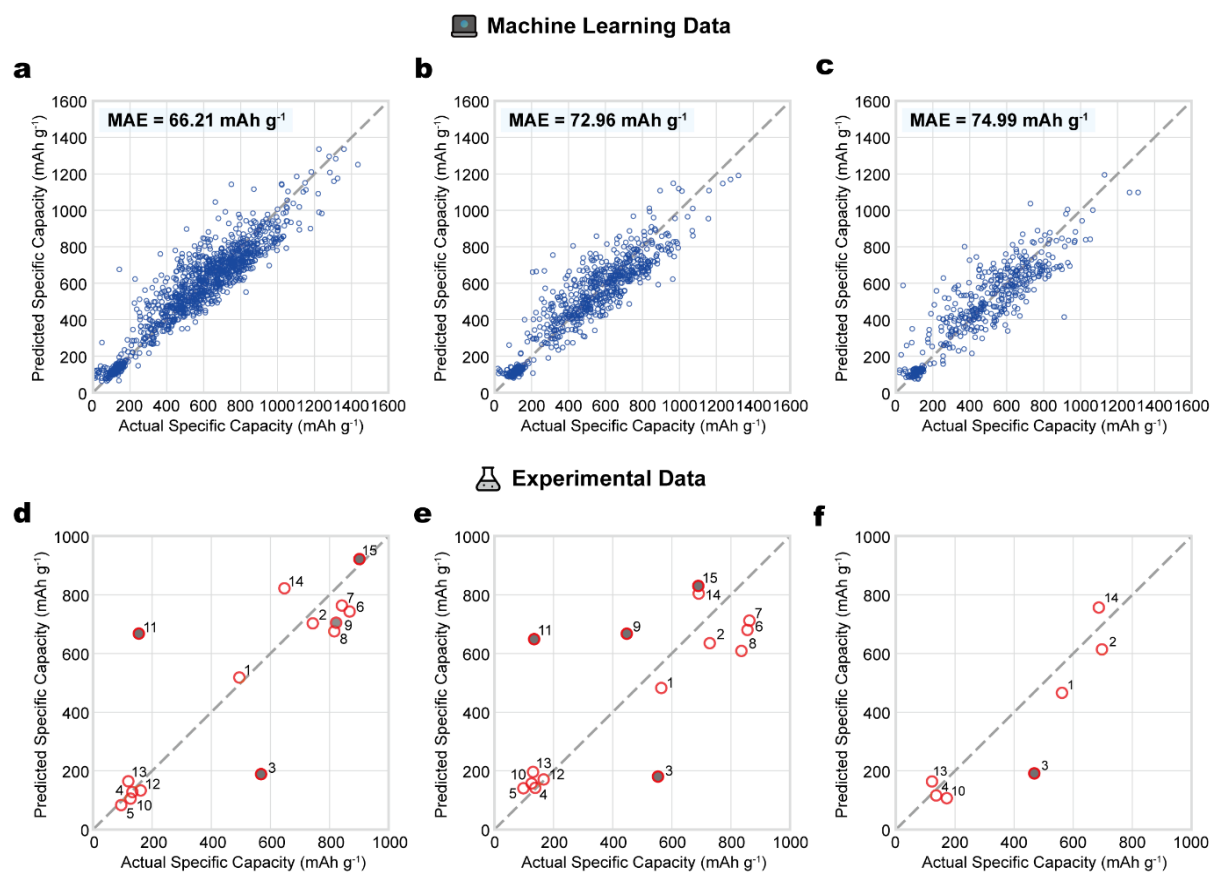

**Fig. 6: Prediction of capacity at target cycle.** Prediction results of gradient boosting regressor model for the capacity of NCM, LFP, sulfur cathode LMBs from **a-c** our database and **d-f** experimental results at the 100th, 200th, and 300th cycles, respectively. Gray-filled red circles indicate battery cells with assembly problems.

Next, we developed machine learning models to predict specific capacities at the 100th, 200th, and 300th cycles (Fig. 6a-c, respectively), utilizing the material compositions and initial capacity values of all types of LMBs, including NCM, LFP, and sulfur cathodes. In addition to the 87 battery component features used in the previous model, a total of 88 features were used, including the initial capacity. For the 100th cycle, a total of 1,252 data points were used; for the 200th cycle, 752 data points; and for the 300th cycle, 498 data points. These data points were divided into training and test sets using a 7:3 ratios. We trained each model using only the cells that had corresponding values at the target cycle. The gradient boosting regressor (GBR) model achieved MAE scores of 66.21, 72.96, and 74.99 mAh g$^{-1}$ at these cycles, respectively (Supplementary Table 12). Note that these values are due to the higher capacity value of LSB (~1600 mAh g$^{-1}$). The R² scores of each model were 0.884, 0.846, and 0.804, indicating good performance.

Unlike the task of predicting initial capacity, high accuracy was obtained even for LSBs by providing the initial capacity information as an input feature. The superior predictive performance of LSBs at specific cycles can be attributed to the model learning the degradation process caused by the shuttle effect. To mitigate the shuttle effect[29], lithium polysulfide migration is typically suppressed through the design of battery components[30-32]. However, similar to the previously mentioned LSB host materials, the absence of a standardized method to encode these materials led us to use only datasets without additional modifications. Consequently, the LSBs in our dataset are expected to exhibit a dominant shuttle effect, which the model appears to have effectively learned. The predictive success at these target cycles, especially for LSBs, is encouraging given that LSB stability characteristics are often assessed at lower C-rates compared to NCM or LFP cathode LMBs. This highlights the potential of our models to early predict long-term cyclability, reducing the need for time-consuming cycle testing.

Additionally, we conducted experiments of cycle tests on 16 different cells with varying cathodes, electrolyte amounts, and C-rates, which were not used as part of the model training data (Supplementary Table 16). As shown in Figure 6d-f, the predictions of our models generally align well with the actual experimental results. However, cell indices 3, 9, 11, and 15(marked with gray-filled red circles) exhibit lower accuracy. Upon examining the cycling profiles of these cells, we observed unstable behavior, such as sudden changes in the slope or prolonged activation periods in the cycle graphs (Supplementary Fig. 19b). This contrasts with other battery cells marked with white-filled red circles (Supplementary Fig. 19a), suggesting that assembly issues may have caused these anomaly behaviors[33]. It is important to note that such anomalies often occur in experiments but are rarely reported, which likely explains their absence in text-mined data. Since the database structure does not account for assembly process variables, these anomalies could not be accurately predicted. Nevertheless, our model performed well when predicting the performance of normally operated cells.

## 3. Prediction of stability at target cycle

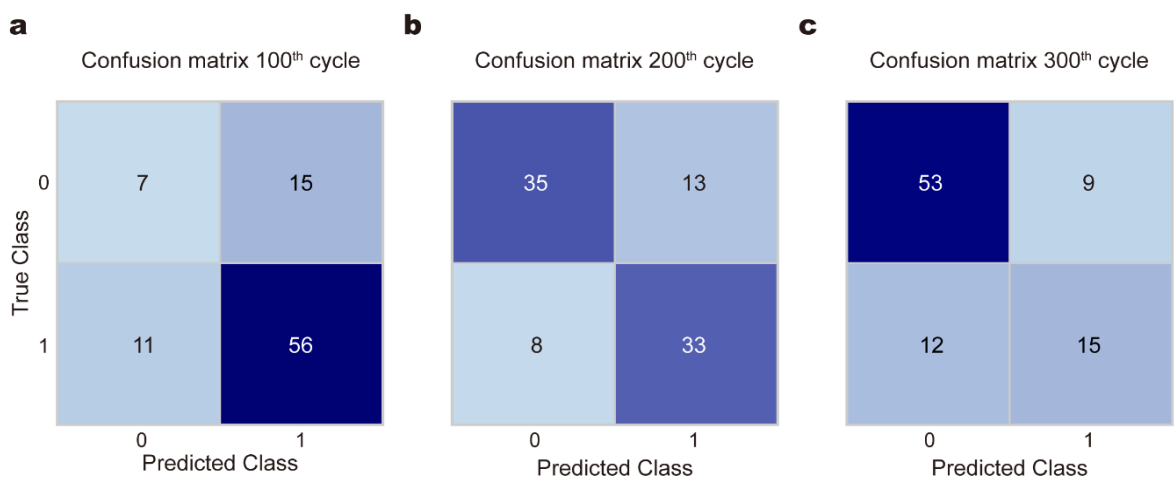

**Fig. 7: Prediction of stability at target cycle.** Confusion matrix of the NCM, LFP batteries' stability prediction model at the **a** 100$^{th}$, **b** 200$^{th}$, and 300$^{th}$ cycles, evaluated on the test dataset.

Finally, we developed models to assess stability of the NCM and LFP cathode batteries at the 100$^{th}$, 200$^{th}$, and 300$^{th}$ cycles, using only material compositions as input features (so no information regarding the initial cycle is given here). Similar to the first task of identifying the prediction of initial capacity, the 87 battery component features were used, with 206 training sets and 89 test sets out of a total of 295 data. Stability was determined by whether the specific capacity fell below 80% of the initial capacity, with cases where the maximum number of cycles was less than the target cycle being classified as unstable. The predictions of stability for LMBs using gradient boosting classifier (GBC) and random forest classifier (RF) models at the 100$^{th}$, 200$^{th}$, and 300$^{th}$ cycles achieved accuracies of 0.816 (GBC), 0.715 (RF), and 0.757 (RF), respectively (Supplementary Table 13).

Researchers usually do not write down or privately record the failed and unstable data. However, these unreported failed data are crucial because the machine learning model the unreported chemical insights can be learned by the machine learning models[34]. Training models with unreported failed data makes it difficult to consider them as high-performing, even if they exhibit high accuracy. As shown in the confusion matrix in Figure 7a, the model predominantly learned to classify data as stable (class 1) due to the abundance of stable data in the test set. Within the NCM and LFP battery dataset of 295 cells, only 74 cells were unstable (class 0) at the $100^{th}$ cycle, while 221 were stable at $100^{th}$ cycle, reflecting an imbalanced data distribution. This phenomenon was also observed in the $300^{th}$ cycle stability prediction model, where 206 cells were unstable and only 89 cells were stable. In contrast, in the case of the $200^{th}$ cycle, the data exhibited a more balanced ratio, with 136 stable cells and 159 unstable cells. This balance resulted in reasonable accuracy, F1 score, and recall values, indicating more appropriate classification performance.

## Discussion

In this research, the ABC platform was developed to automatically extract multi modal battery information from research articles and perform data-driven analysis using an early prediction machine learning model. By integrating an automatic graph mining tool (MatGD) and a large language model (LLM), our ABC model successfully extracted refined information from 8,074 filtered battery cells. We developed machine learning models to explore the relationships between battery performance and material properties using the compiled database. These models made significant predictions, including: 1) the initial capacity of NCM and LFP cathode LMBs, 2) the capacity at the $100^{th}$, $200^{th}$, and $300^{th}$ cycles for NCM, LFP, and sulfur cathode LMBs, and 3) the stability at the $100^{th}$, $200^{th}$, and $300^{th}$ cycles for NCM and LFP cathode LMBs. Our innovative multi modal data-driven machine learning techniques not only improve the efficiency and accuracy of device data mining but also demonstrate versatility for application to other devices and materials. Furthermore, these techniques have the potential to be combined with reinforcement learning and autonomous experimentation, thereby advancing the discovery of new materials across a wide range of research areas

**Methods**

**Journal Paper Parsing and Crawling**

The parsing agent focused primarily on scientific journals from the Elsevier publisher. Out of a total of 7,342 LMB-related papers, it secured 6,444 papers, excluding 898 papers that weren't related to our targets: NCM, LFP, sulfur cathode, and liquid electrolyte. During this process, the parsing agent identified papers containing LMB-related terms in their titles, abstracts, and keywords, gathering a collection of papers that were highly relevant to our research objectives. To preserve the integrity of our research and comply with copyright laws, the parsing agent downloaded the journal papers in XML format, following the data usage policies set by Elsevier journals. Additionally, it utilized the Elsevier Scopus API for efficient data filtering.

**Prompt Engineering**

LLM is a trained artificial intelligence (AI) system capable of understanding the context in which language is used and completing natural language processing tasks by applying learned patterns. Prompt engineering has emerged as a crucial strategy for tailoring language models to specific objectives and enhancing their performance[35,36]. This process involves the creation of advanced prompts designed to extract highly accurate battery information from research articles. We have developed an innovative set of prompts to decompose and reconstruct battery components, which are integrated with the LLM via Python code.

All prompts are constructed using an objective-question-explanation-exemplars structure, enabling precise problem-solving and ensuring the output format meets specific requirements. During all extraction stages, the LLM uses detailed explanations in the prompts to gain a comprehensive understanding of the battery device and accurately identify the relevant classes, properties, or materials from the target paragraph. In the merging stage, the process involves using text mining and graph mining results to identify battery cell information within the cycle graphs. A key aspect of our prompt engineering is that the outcomes from each sequential stage are passed on to the next, facilitating the rapid and efficient generation of progressive mining results. Each prompt is reinforced with two to four examples to enhance the accuracy of the LLM in generating responses.

**Graph Mining**

MatGD is designed to identify and eliminate objects such as text, legends, and arrows from graphs, based on the YOLOv8 model architecture. It employs the DBSCAN algorithm to segregate the remaining data lines based on their RGB color vector. However, this approach imposes limitations on MatGD's capability to digitize certain types of graphs accurately.

For example, inaccuracies occur when a marker from a legend, which should be removed as part of the graph objects, is not deleted and is mistakenly extracted as a data point. Additionally, challenges arise when different battery cells are depicted in the same color but are differentiated by scatter point shapes, or when the RGB values of the original graph's data lines are similar, leading to improper separation. As a result, such graphs are excluded to ensure that only accurate data is extracted.

To eliminate Coulombic efficiency (CE) data and isolate the cyclability data, specific identification methods were used. When the label for CE was present in the graph legend and detected through text recognition, the corresponding CE data were removed directly. In cases where the CE label was absent, tick labels or axis labels on the right side of the graph box were used as indicators to determine the presence of CE data. A threshold was set based on the observation that CE values typically stabilize around 100% after a few cycles. By using this threshold, CE data could be effectively identified and eliminated, allowing for the isolation of the capacity data line.

**Feature Standardization**

Despite the identical chemical composition or materials, authors in various studies often describe them differently, making a standardized process essential During this standardization process, materials amenable to SMILES format conversion, including binders, conductive materials, electrolyte salts, electrolyte solvents, and separators, were transformed.

Subsequently, the units of extracted features, including the thickness of lithium anodes, temperature, current density, and loading of active materials, were standardized using rule-based Python code to harmonize these units and convert associated values. This unit standardization was also applied to the concentration of electrolyte salt and the ratio of electrolyte solvent. For consistent representation of electrolyte concentrations, different metrics like molarity, molality, and mass percentage were converted, with additional information such as molecular weight and density extracted using SMILES.

**Machine learning-input data**

For our machine learning analyses, we utilized the merged database previously described. The active materials, conductive additives, binders, salts, and solvent materials were all converted into SMILES notation and encoded into one-hot vectors, with the presence or absence of these materials indicated using binary values (1 for presence, 0 for absence). The electrolyte salts and solvents were represented as vectors, expressed in terms of their molar concentrations and volume ratios. Our study only included pure Celgard separators in the dataset due to the challenges of representing modified separators, such as doped or mixed Celgard, in SMILES format. Additionally, we restricted our analysis to battery cells operated at room temperature (RT cells), resulting in a dataset that includes 1,736 battery cell data points.

**Materials and electrochemical measurements**

Lithium metal foils (250μm), and carbon coated Al foils were purchased from MTI corporation. DME, EC, DEC, DMC dehydrated solvent and Lithium bis(trifluoromethanesulfonyl)imide (LiTFSI, 99.99%), Lithium bis(fluorosulfonyl)imide

(LiFSI, >99.9%), Lithium difluoro(oxalato)borate (LiDFOB), Lithium nitrate (LiNO$_3$, 99.9%), and Lithium hexafluorophosphate (LiPF$_6$, 99.9%) salts were purchased from Sigma-Aldrich. Lithium bis(pentafluoroethanesulfonyl)imide (LiBETI, 98.0%) was purchased from Tokyo Chemical Industry Co. All electrolytes used in each machine learning model were prepared by dissolving specific amounts of salt and solvent in an Ar-filled glove box (O$_2$ < 0.1ppm, H$_2$O < 0.1ppm). NCM523 [(3.4 mAh/cm$^2$) NCM523 : Super P C65 : Polyvinylidene fluoride (PVDF) = 94 : 3 : 3], NCM622 [(1.56 mAh/cm$^2$) NCM622 : Super P C67 : PVDF = 85 : 10 : 5], and NCM811 [(2.2 mAh/cm$^2$) NCM811 : Super P C65 : PVDF = 96 : 2 : 2] cathodes were purchased from Welcos Ltd. Areal loadings of active materials for NCM 523, 622, and 811 were 21.5mg/cm$^2$, 11.9mg/cm$^2$ and 12mg/cm$^2$ respectively. LFP cathode (2.5mAh/cm$^2$) with 17.24mg/cm$^2$ was purchased from Welcos Ltd. To prepare sulfur cathodes, melt diffusion was employed using 70wt% Sulfur (Sigma-Aldrich) and 30wt% carbon host, either Super P carbon (SP, Thermo Fisher) or carbon nanotube (CNT, Carbon Nano-material Technology Co.), at 155°C for 12 hours. Subsequently, we mixed sulfur/carbon composite with SP conductive additive and PVDF (Mw ~534,000, Sigma-Aldrich) binder in a specific mass ratio. This mixture was dispersed in N-methyl-2-pyttolidone solvent (Sigma-Aldrich) to form a slurry. We then bar-coated (Welcos Ltd.) slurry onto carbon coated Al foil, which was dried in a vacuum oven at 60°C for 8 hours.

All coin-type cells (CR2032) were assembled with a 12mm diameter of cathode, PP separator (Celgard 2400, Welcos Ltd.) and a 16mm diameter of Li anode in an Ar-filled glove box (O$_2$ < 0.1ppm, H$_2$O < 0.1ppm). The amount of injected electrolyte was adjusted to a specific E/A ratio according to the experimental conditions. The galvanostatic charge/discharge cycle tests were conducted within a voltage window of 3.0-4.3V (NCM), 2.8-4.3V (LFP), and 1.7-2.8V (Sulfur) at room temperature using a WBCS3000L (WonATech Ltd.).

## Code availability

The ABC model is available at https://github.com/skyljw0714/Automatic-battery-collector .

## Acknowledgements

This work is supported by Basic Science Research Program through the National Research Foundation of Korea(NRF) funded by the Ministry of Education (RS-2024-00411362). S. Park was supported by the National Research Foundation of Korea (NRF) grant funded by the Korea government (MSIT) (No. 2019R1A5A8080326).

## Author contributions

J.L and J.W contributed equally to this work. J.L and J.W developed ABC model and wrote the manuscript with J.K, S.P, H.K. Assistance in the experimental validation was provided by S.K, C.P, H.P. The manuscript was written through the contributions of all authors. All authors have given approval for the final version of the manuscript.

## Conflicts of Interests

There are no conflicts to declare.